\documentclass[runningheads]{llncs}

\usepackage{tikz}
\usepackage{amsmath,amssymb} %
\usepackage{color}
\usepackage[accsupp]{axessibility}  
\usepackage{algorithm,multirow}
\usepackage{graphicx}
\usepackage{amssymb}
\usepackage{booktabs}
\usepackage{algorithm,multirow}
\usepackage{amsfonts,bm,,algorithm, algpseudocode}
\usepackage{blindtext}
\usepackage{subfigure}

\newcommand\B{\rule[-1.2ex]{0pt}{0pt}} %
\newcommand\T{\rule{0pt}{2.6ex}}       %

\newif\ifdraft
\drafttrue

\newcommand{\bq}{\mathbf{q}}

\newcommand{\bI}{\mathbf{I}}

\newcommand{\bh}{\mathbf{h}}
\newcommand{\bM}{\mathbf{M}}

\newcommand{\bB}{\mathbf{B}}

\definecolor{orange}{rgb}{1,0.5,0}

\newcommand{\ms}[1]{\ifdraft {\color{orange}{#1}} \else {#1}\fi}
\newcommand{\kn}[1]{\ifdraft {\color{blue}{#1}} \else {#1}\fi}

\newcommand{\KN}[1]{\ifdraft {\color{red}{\textbf{KN: #1}}}\else {}\fi}

\newcommand{\comment}[1]{}

\iftrue %

\else %

\fi

\begin{document}
	
\pagestyle{headings}
\mainmatter

\title{{Understanding Pose and Appearance Disentanglement in 3D Human Pose Estimation}}

\author{%
	{Krishna Kanth Nakka $^{1}$ and \quad Mathieu Salzmann $^{1,2}$}
}
\titlerunning{Pose and Appearance Disentanglement}
\authorrunning{KK Nakka and M Salzmann}
\institute{
    {\small $^1$ EPFL CVLab, \quad $^2$ ClearSpace SA} \\
    \email{{\tt \small krishkanth.92@gmail.com and \quad mathieu.salzmann@epfl.ch}}
}

\maketitle

\begin{abstract}
As 3D human pose estimation can now be achieved with very high accuracy in the supervised learning scenario, tackling the case where 3D pose annotations are not available has received increasing attention. In particular, several methods have proposed to learn image representations in a self-supervised fashion so as to disentangle the appearance information from the pose one. The methods then only need a small amount of supervised data to train a pose regressor using the pose-related latent vector as input, as it should be free of appearance information. In this paper, we carry out in-depth analysis to understand to what degree the state-of-the-art disentangled representation learning methods truly separate the appearance information from the pose one. {First, we study disentanglement from the perspective of the self-supervised network, via diverse image synthesis experiments. 
Second, we investigate disentanglement with respect to the 3D pose regressor following an adversarial attack perspective. Specifically, we design an adversarial strategy focusing on generating natural appearance changes of the subject, and against which we could expect a disentangled network to be robust. Altogether, our analyses show that disentanglement in the three state-of-the-art disentangled representation learning frameworks if far from complete, and that their pose codes contain significant appearance information.}
We believe that our approach provides a valuable testbed to evaluate the degree of disentanglement of pose from appearance in self-supervised 3D human pose estimation.

\end{abstract}

\vspace*{-1.1cm}
\section{Introduction}\label{sec:Intro}

\begin{figure}[t]
				\vspace{-1.3em}
				\centering
				\begin{minipage}[c]{.45\textwidth}
					\begin{flushleft}
						\caption{ {\bf Disentanglement-based Representation Learning.} Given a reference frame and another frame from either a different view or a different time instant, an encoder learns a representation separated into two components,  appearance and pose, in a self-supervised fashion.  A  pose regressor is then trained using limited annotated data to map the latent pose vector to a 3D human pose.
							   }   \label{fig:teaser}
					\end{flushleft}
				\end{minipage}%
				\hspace{0.1em}
				\begin{minipage}[c]{.5\textwidth}
					\centering
					\begin{flushright}
						\includegraphics[width=1\textwidth,height=0.6\textwidth]{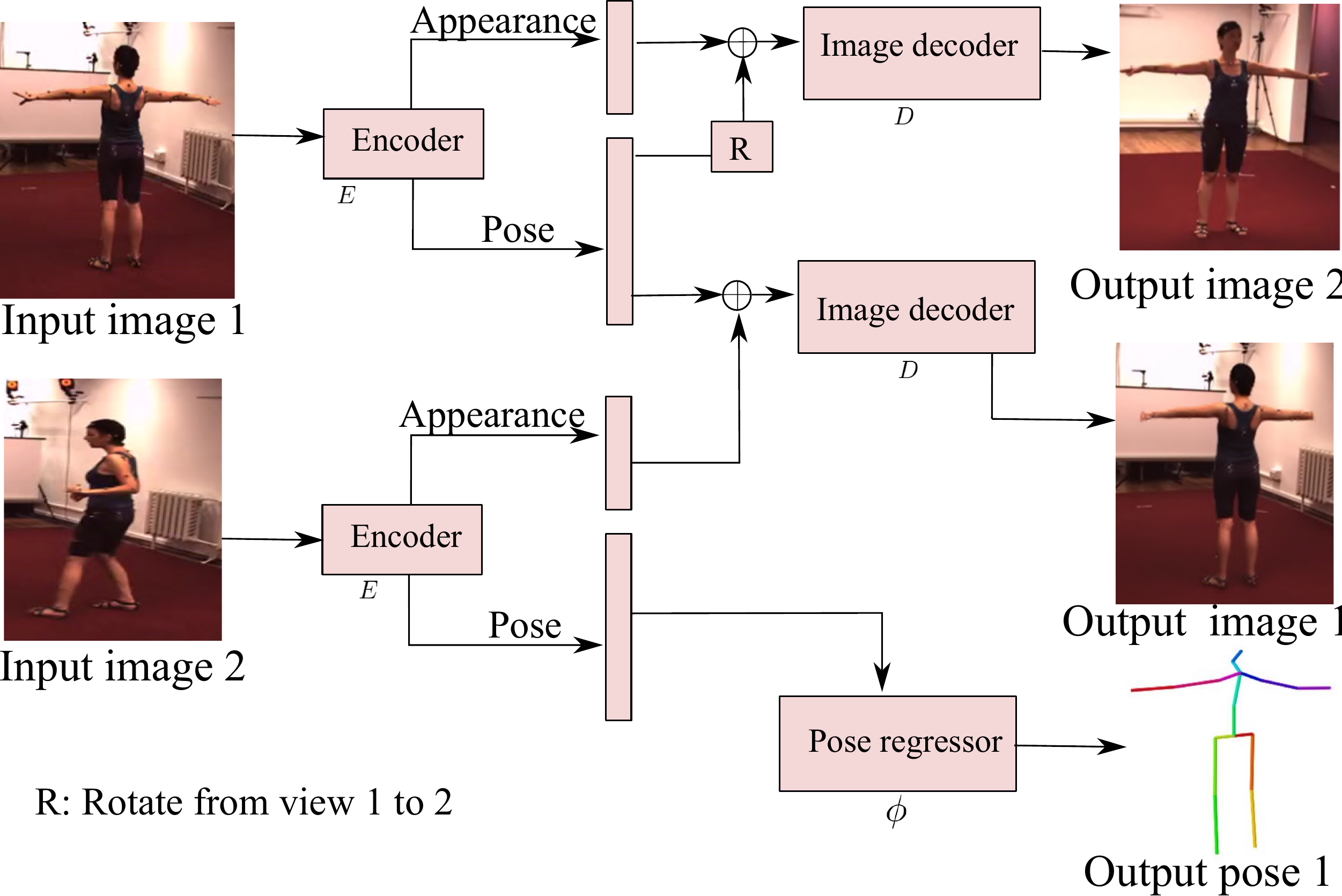}
					\end{flushright}
				\end{minipage}
			\end{figure}

Monocular 3D human pose estimation has been at the heart of computer vision research for decades, and tremendous results can now be achieved in the supervised learning setting~\cite{mehta2017monocular,ionescu2014iterated,joo2015panoptic,pavlakos2017coarse,tome2017lifting,popa2017deep,mehta2017vnect,tekin2017learning,pavlakos2017harvesting,rogez2017lcr,martinez2017simple}. Unfortunately, obtaining 3D pose annotations for real images remains very expensive, particularly in the wild. As such, self-supervised learning approaches have received an increasing attention in the past few years~\cite{rhodin2018,NSD,CSSL,DRNet}. One of the common factors across all these methods is their aim to learn a latent representation of the image that disentangles the person's pose from their appearance. In practice, {as shown in Figure~\ref{fig:teaser},} this has been achieved by leveraging access to either multiple views~\cite{NSD,rhodin2018} or video sequences~\cite{DRNet,CSSL} during training. In either case, one then only needs access to a small amount of supervised data to effectively train a pose regressor from the pose-related portion of the latent code to the actual 3D pose, because this portion of the latent code should in theory contain only pose-relevant information.

Despite the impressive progress of these self-supervised 3D human pose estimation methods, several fundamental questions about their learnt representations remain unanswered.  For example, to what extent are the pose and appearance latent vectors disentangled?  Do these two representations contain truly complementary information, or do they share some signal? How do the different sources of self-supervision, i.e., multiple views or temporal information, affect the disentanglement of these  representations?

In this paper, we seek to provide a deeper understanding of such disentangled representations by analyzing the resulting latent spaces in two ways.  First, we study the disentanglement of the latent pose and appearance vectors with respect to the self-supervised representation learning network. In this context, we analyze both the images synthesized by altering the appearance codes in different ways, and the 
{influence on pose and appearance of different channels in the latent pose codes.}
Second, we investigate the disentanglement with respect to the supervised 3D pose regressor.
To this end, we follow an adversarial attack strategy, aiming to modify the subject's appearance so as to affect the regressed 3D pose. However, instead of exploiting a standard adversarial attack technique~\cite{madry2018towards,kurakin2016adversarial,goodfellow2014explaining}, against which disentangled pose networks were never meant to be robust, we design a dedicated framework that should be much more favorable to such networks. Specifically, we seek to alter only the latent appearance vector so as to affect the 3D pose regressed from the latent pose vector extracted from the image synthesized using the modified appearance vector with the original pose one.

Our  experiments on the state-of-the-art disentangled representation learning frameworks, NSD~\cite{NSD}, CSSL~\cite{CSSL} and DRNet~\cite{DRNet},  evidence that, across the board, \emph{disentanglement is not complete and the pose codes of these frameworks contain appearance information.} {Our work provides the tool to study the effectiveness of different disentanglement-based training strategies and will serve as a valuable testbed to analyze the extent of disentanglement in future frameworks.}

\noindent{\bf Contributions.} To summarize, our contributions are twofold. (1) We systematically analyze the latent pose and appearance representations in several representative disentangled networks.  Our experiments lead to an interesting finding that the latent pose vectors contain almost all of the subject's appearance information.
	(2)  We introduce an adversarial strategy to understand the disentanglement of 3D pose from natural appearance changes.  
Our code and trained networks will be made publicly available upon acceptance.

\vspace*{-0.3cm}
\section{Related Work}\label{sec:relatedwork}

\paragraph{\bf Disentanglement-based 3D Human Pose Estimation.}

Disentangling pose and appearance in 3D pose estimation was first proposed in DRNet~\cite{DRNet}, where a discriminator was employed to distinguish if the time-varying features from two images represented the same subject or not. Furthermore, the distance between the time-invariant, i.e., appearance, component of one subject at two different time instants was minimized, and the time-varying pose features were encouraged to be indistinguishable {across subjects}, 
thereby ensuring that appearance information did not leaked into the pose features. In~\cite{NSD,rhodin2018}, disentanglement was achieved via the use of multiple views during training, leveraging the intuition that, for one subject, the pose features extracted from one view and rotated to a different view at the same time instant should be the same as those directly extracted from that view, and the appearance features at different time instants should be similar so as not to contain pose information. More recently, \cite{CSSL} designed a contrastive loss to force the latent appearance features in temporally-distant frames to remain close while encouraging the pose features to be different from each other.  All these methods learn the disentangled representation from unsupervised data, and then train a shallow regressor to predict 3D pose from the pose latent vector using a limited amount of pose labels. In this work, we study how disentangled the appearance and pose features extracted by these methods truly are. To this end, we provide analyses based on {diverse image synthesis experiments
and on adversarial attacks.}

\noindent{\bf Adversarial Attacks.}  Deep neural networks were first shown to be vulnerable to adversarial examples in~\cite{szegedy2013intriguing}. Following this, several attacks have been proposed, using either gradient-based approaches~\cite{goodfellow2014explaining,kurakin2016adversarial} or optimization-based techniques~\cite{carlini2017towards,jsma,moosavi2016deepfool,croce2019sparse,dong2019evading}. {To study the disentanglement of pose and appearance in 3D human pose estimation, we seek to analyze if appearance changes can affect the regressed 3D pose. In principle, we could use any of the above-mentioned attack strategy to do this. However, they offer no control on the generated perturbations, and thus could potentially incorporate structures that truly suggest a different pose. In other words, the disentangled networks cannot be expected to be robust to such attacks. Therefore, we design an attack strategy to which they can be expected to be robust. Specifically, we synthesize an image by modifying only the appearance code of the network of interest, and show that the 3D pose regressed from that image will typically differ from the original one.}
Our attacks can be thought of as inconspicuous ones, as the generated image looks natural, with only appearance changes to the subject. Other works~\cite{zhao2018generating,joshi2019semantic,qiu2020semanticadv,songconstructing,bhattad2019unrestricted,sharif2017adversarial} have designed strategies to generate realistic adversarial images, typically focusing on face recognition datasets and using GANs~\cite{goodfellow2014generative,mirza2014conditional,arjovsky2017wasserstein}. Our approach nonetheless fundamentally differs from those in both methodology and context; our main goal is not to attack disentangled 3D human pose networks but to study their level of disentanglement. Therefore, we design an attack strategy that is most favorable for these networks, and against which they can be expected to be naturally robust.

{
\noindent{\bf Measuring Disentanglement.} In other contexts than human pose estimation, several works have proposed metrics to quantify the degree of disentanglement of latent vectors~\cite{eastwood2018framework,do2019theory,liu2021measuring}. These methods are of course also applicable to the self-supervised learning frameworks that we will analyze, and we will report these metrics in our experiments. However, these metrics do not provide any understanding of where disentanglement fails. This is what we achieve with our diverse analyses. 
}

\vspace*{-0.4cm}

\section{Disentangled Human Pose Estimation Networks}
\label{sec:disentangled}

Given an image as input, 3D human pose estimation aims to predict the 3D positions of $J$ body joints, such as the wrists, elbows, and shoulders. When no annotations are available for the training images, an increasingly popular approach consists of learning a latent representation that disentangles appearance from pose in a self-supervised fashion. Here, we review disentanglement-based 3D human pose estimation frameworks that we will analyze in Sections~\ref{sec:synthesis} and~\ref{sec:attacks}.

Existing disentanglement-based frameworks essentially all follow the same initial steps. The input image $\bI$ is first passed through a spatial transformer network  $\mathcal{S}$  to extract the bounding box corresponding to the human subject. An encoder $E$ then takes the cropped bounding box $\bI_c$ as input and outputs a latent vector $\bh$ comprising two components, that is $E : \bI_c \rightarrow [\bh_a, \bh_p]$. The first component, $\bh_a$, aims to encode the subject's appearance while the second, $\bh_p$, should represent the subject's pose. The networks are trained without any 3D pose annotations, and thus
supervision is achieved via image reconstruction. Specifically, a decoder $D$ takes the complete the latent vector $\bh$ as input and and outputs a reconstructed version of the cropped image $\tilde{\bI}_c$, with an additional mask $\bM$ corresponding to the subject's silhouette. The cropped image is further merged with a pre-computed background image $\bB$ to obtain the final reconstructed input image $\tilde{\bI}$.  

The main difference between existing frameworks lies in the way they encourage the disentanglement of pose and appearance. Specifically, the different frameworks train the encoder $E$ and decoder $D$ as follows:

\noindent{\bf NSD~\cite{NSD}.} The neural scene decomposition (NSD) approach leverages the availability of multiple views during training. Given a pair of images from two views at time $t$, NSD passes one image to the encoder to obtain an appearance vector $\bh_a^t$ and a pose vector $\bh_p^t$. The pose vector $\bh_p^t$, shaped as a 3D point cloud, is rotated to the second view using the ground-truth camera calibration between the two views to obtain a transformed pose vector $\bh_{p, r}^t$.   Furthermore, to factor out appearance from pose, NSD replaces the appearance vector $\bh_a^t$ by an appearance vector $\bh_a^{t_1}$ of the same subject at a different time instant $t_1$. The decoder $D$ then takes as input $\bh=[\bh_{p,r}, \bh_a^{t_1}]$ and aims to reconstruct the image from the second view at time $t$.

\noindent{\bf CSSL~\cite{CSSL}.} Instead of using multiple views, contrastive self-supervised learning (CSSL) exploits temporal information from videos to learn a latent representation of pose and appearance. To achieve disentanglement, CSSL encourages the distance between the latent pose vectors $\bh_p^{t_1}$ and $\bh_p^{t_2}$ of two frames, $t_1$ and $t_2$, to reflect their temporal distance. Furthermore, similarly to NSD, CSSL swaps the appearance vectors $\bh_a^{t_1}$ and $\bh_a^{t_2}$ of the two video frames when performing image reconstruction so as to force them to learn time-invariant information, thus encoding appearance. 

\noindent{\bf DRNet~\cite{DRNet}.} The disentangled representation network (DRNet) uses a similar strategy to that of CSSL, consisting of randomly choosing two temporal frames, $t_1$ and $t_2$, from a video.  However, DRNet aims to achieve disentanglement in two ways: (1) By minimizing the distance between the two appearance vectors $\bh_a^{t_1}$ and $\bh_a^{t_2}$; and (2) by exploiting an adversarial network to make the pose vector $\bh_p$ independent of the subject's appearance. Specifically, this is achieved by training the additional discriminator to output the subject's identity given the pose vector as input, and training the encoder $E$ in an adversarial fashion to fool the discriminator.

Once trained on a large corpus of unannotated images in a self-supervised manner, the frameworks discussed above employ a 2 layer pose regressor $\phi: \bh_p \rightarrow \bq$ to predict the 3D pose $\bq$ from the latent pose vector $\bh_p$. This pose regressor is trained with a small amount of supervised data, while freezing the weights of the encoder. \kn{Due to space limitations, we provide additional details about training in the supplementary material.}

\section{Disentanglement w.r.t. the Self-Supervised Network}

In this section, we study the disentanglement of pose and appearance within the self-supervised representation learning network itself. To this end, we first analyze the impact of the latent appearance vector on the images synthesized by the network's decoder. {We then turn to investigating the influence on pose and appearance of different channels in the latent pose vector.} %

\label{sec:synthesis}

\vspace*{-0.3cm}
\subsection{Effect of the Appearance Vector on Synthesized Images}

\begin{figure}[!t]
    \begin{center}
    \addtolength{\tabcolsep}{-5pt}  %
    \begin{tabular}{cccc}
    \includegraphics[width=0.8\linewidth, height=10cm]{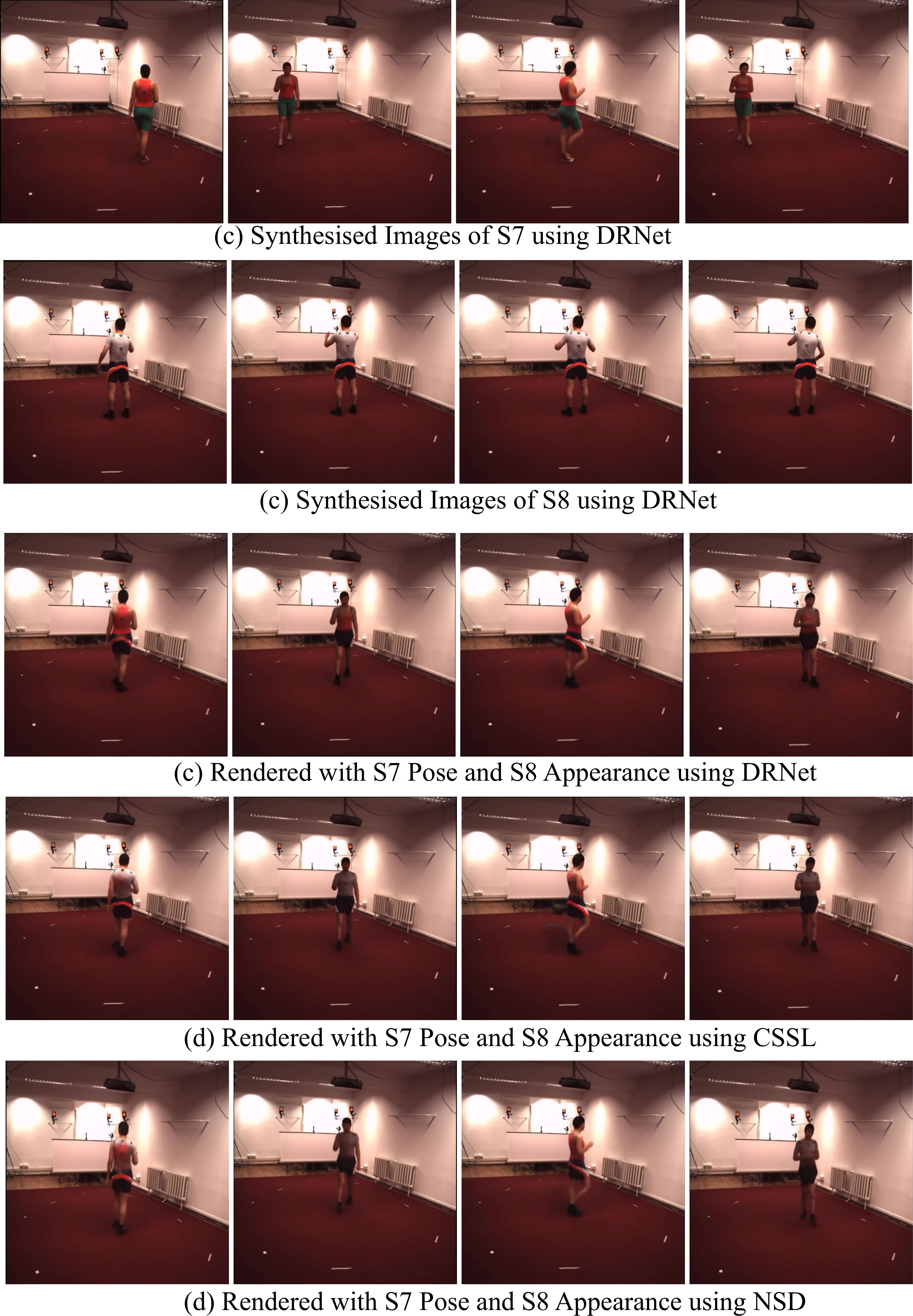}
    \end{tabular}
    \addtolength{\tabcolsep}{5pt}  %
    \end{center}
    \vspace{-6mm}
    \caption{ {\bf Synthesizing novel images.} We take the pose information from S7 (first row) and the appearance information from S8 (second row) and synthesize novel images in the third, fourth and fifth rows using DRNet, CSSL and NSD. The synthesized images retain some appearance information (red shirt) of S7 although we only use S7's pose code in the synthesis.}
    \label{fig:analysis1_main}
\end{figure}

Our first analysis consists of visualizing the images generated by the network's decoder. In particular, we leverage the intuition that, if the pose and appearance vectors were disentangled, altering the appearance vector while keeping the pose one fixed should yield images with a different subject's appearance but the same pose. We investigate this via the two strategies discussed below.

First, we synthesize novel images by mixing the appearance and pose information from two subjects, S8 and S7. The top two rows of Figure~\ref{fig:analysis1_main} show the images synthesized with DRNet\footnote{{Similar images for the other networks are provided in the supplementary material.}} \emph{without mixing the appearance vectors; these images look similar to the original ones, depicting two clearly different subject's appearances.}  By contrast, the images in the third to fifth row of the figure, obtained by using S7's pose vector and S8's appearance one, still contain appearance information of S7. This is particularly the case for the images synthesized using DRNet and NSD, in which the subject's shirt has taken the red color of that of S7, although we use only S7's pose code in the synthesis process. CSSL is less subject to such failures, but they nonetheless occur in some cases, such as in the third and fourth columns.

\begin{figure}[!t]
    \begin{center}
    \addtolength{\tabcolsep}{-5pt}  %
    \begin{tabular}{cccc}
    \includegraphics[width=\linewidth]{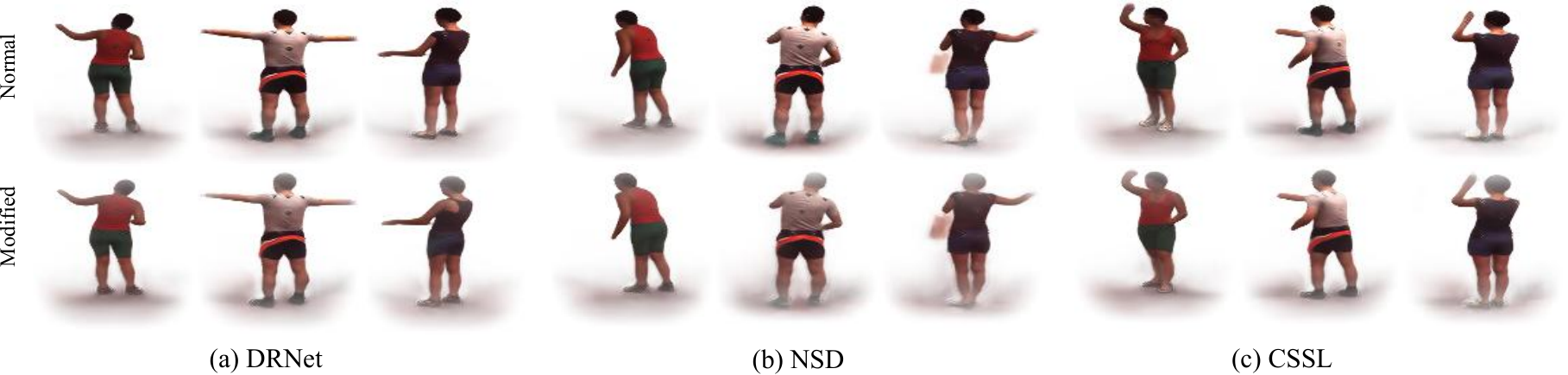}
    \end{tabular}

    \addtolength{\tabcolsep}{5pt}  %
    \end{center}
    \vspace{-6mm}
   \caption{ {\bf Replacing the appearance code with a  fixed zero vector.} In the first row, we show the original synthesized images for four subjects. In the second row, we set the values in the appearance vector to zero and use the same pose vectors as in the first row. Despite using a similar zero appearance vector, the outputs do not appear similar in content and instead retain almost all the appearance information of the original images.
   }
    \label{fig:analysis2_main}
\end{figure}

As a second experiment, we replace the appearance vector with a zero vector. We then combine this zero appearance vector with the pose vector obtained from the original image shown in the first row of Figure~\ref{fig:analysis2_main}. As can be seen from the second row, even though we use the same zero appearance vector to generate images of different subjects, the synthesized images retain almost all the appearance information of the original images, except near the head region. 

\begin{figure}[!t]
    \begin{center}
    \addtolength{\tabcolsep}{-5pt}  %
    \begin{tabular}{cccc}
    \includegraphics[width=\linewidth]{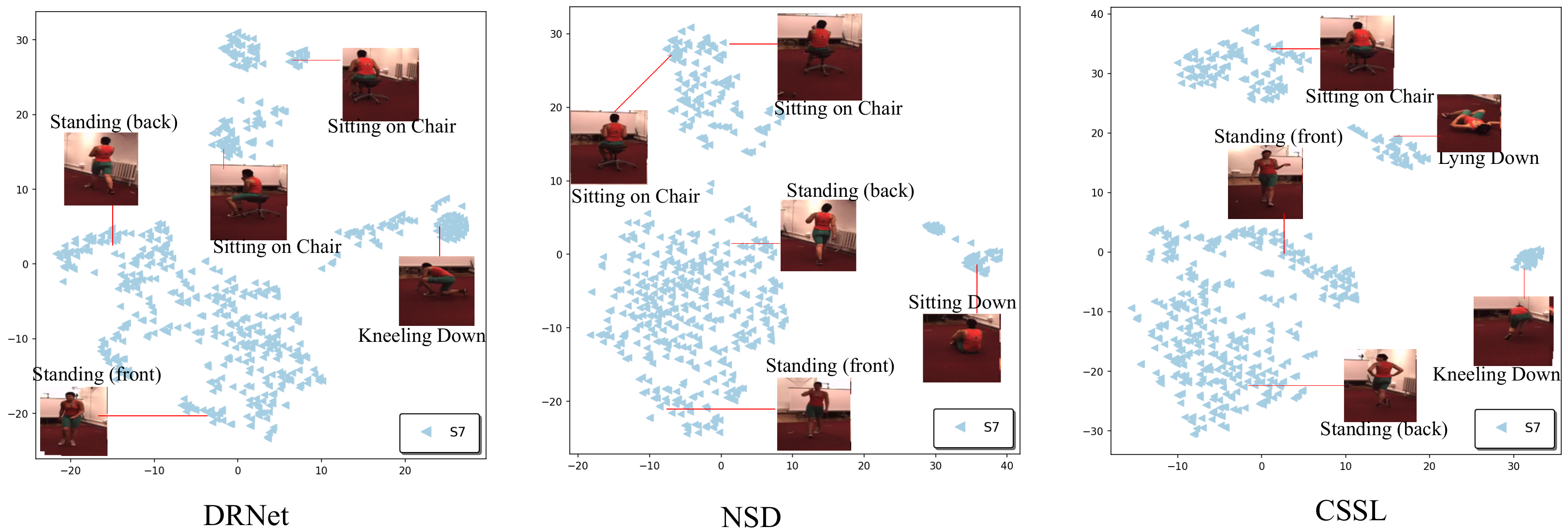}
    \end{tabular}
    \addtolength{\tabcolsep}{5pt}  %
    \end{center}
    \caption{ 	
    	{	{\bf tSNE visualization of  appearance codes.} The appearance codes of images from same subject S7 are clustered according to the action performed by the subject. This indicates that the appearance code still contains  information about the pose. Best viewed in color and zoomed in.} 
    }
    \label{fig:tsne_app}
\end{figure}

Both of these experiments evidence that the pose code contains a significant amount of appearance information and that the disentanglement is thus not complete.  Nevertheless, both experiments also show that modifying the appearance code indeed does not impact the subject's pose in the synthesized image. {To further verify whether the appearance codes are truly free of pose information, we visualize the appearance codes of all images of a   S7 using t-SNE in Figure~\ref{fig:tsne_app}. The resulting plot shows nicely-separated clusters, which can be observed to correspond to action categories. This suggests that, although modifying the appearance code does not visually change the subject's pose in the synthesized images, the appearance codes still contain information about the subject activity, and thus about their pose.}

\subsection{Effect of the Pose Vector on Synthesized Images}

\begin{figure*} [t] \centering\setlength{\tabcolsep}{0cm}
				\resizebox{0.4\textwidth}{!}{
	\subfigure[ { \small Two images with similar poses but different appearances.} ]
	{
		\begin{tabular}{@{}cccc@{}}
    \includegraphics[width=0.45\linewidth, height=2.5cm]{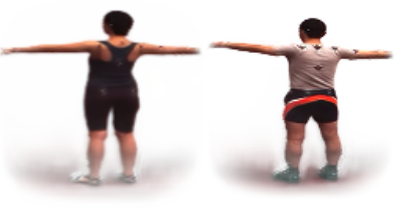}
	\end{tabular}\label{fig:tsne_strictBB}} 
}
				\resizebox{0.35\textwidth}{!}{
	\subfigure[ {Absolute difference between the pose codes sorted by magnitude.} ]
	{
		\begin{tabular}{@{}cccc@{}}
        \includegraphics[width=0.4\linewidth, height=2.5cm]{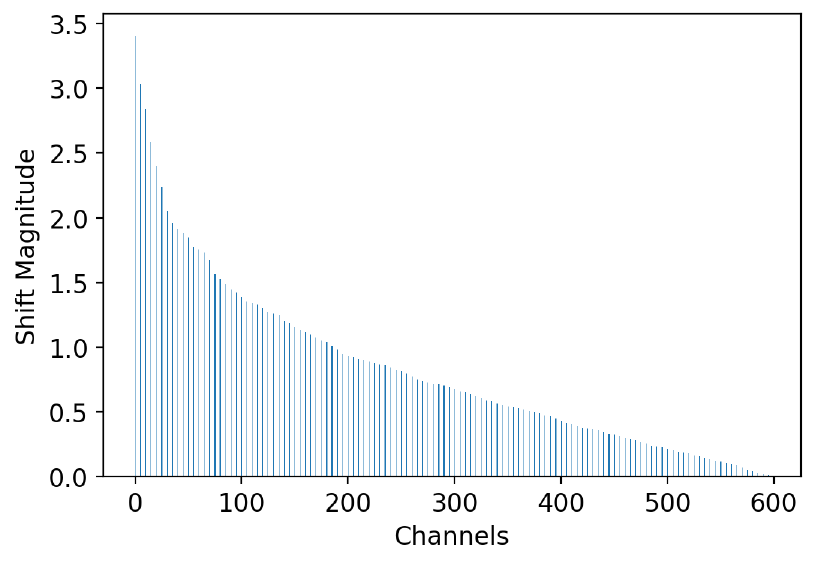}
		\end{tabular}\label{fig:tsne_crossdomainBB}} }
	\vspace*{-0.3cm}
	    \caption{ {{\bf Detecting appearance channels in the pose latent vector.} We take images depicting different subjects in a similar pose, for which we could expect the pose codes to be close. However, as shown on the right, the latent pose vectors obtained by NSD contain channels with large differences, likely to encode appearance information.}}
    \label{fig:analysis3_additional}
\end{figure*}

\begin{figure}[!t]
    \begin{center}
    \addtolength{\tabcolsep}{-5pt}  %
    \begin{tabular}{cccc}
    \includegraphics[width=\linewidth]{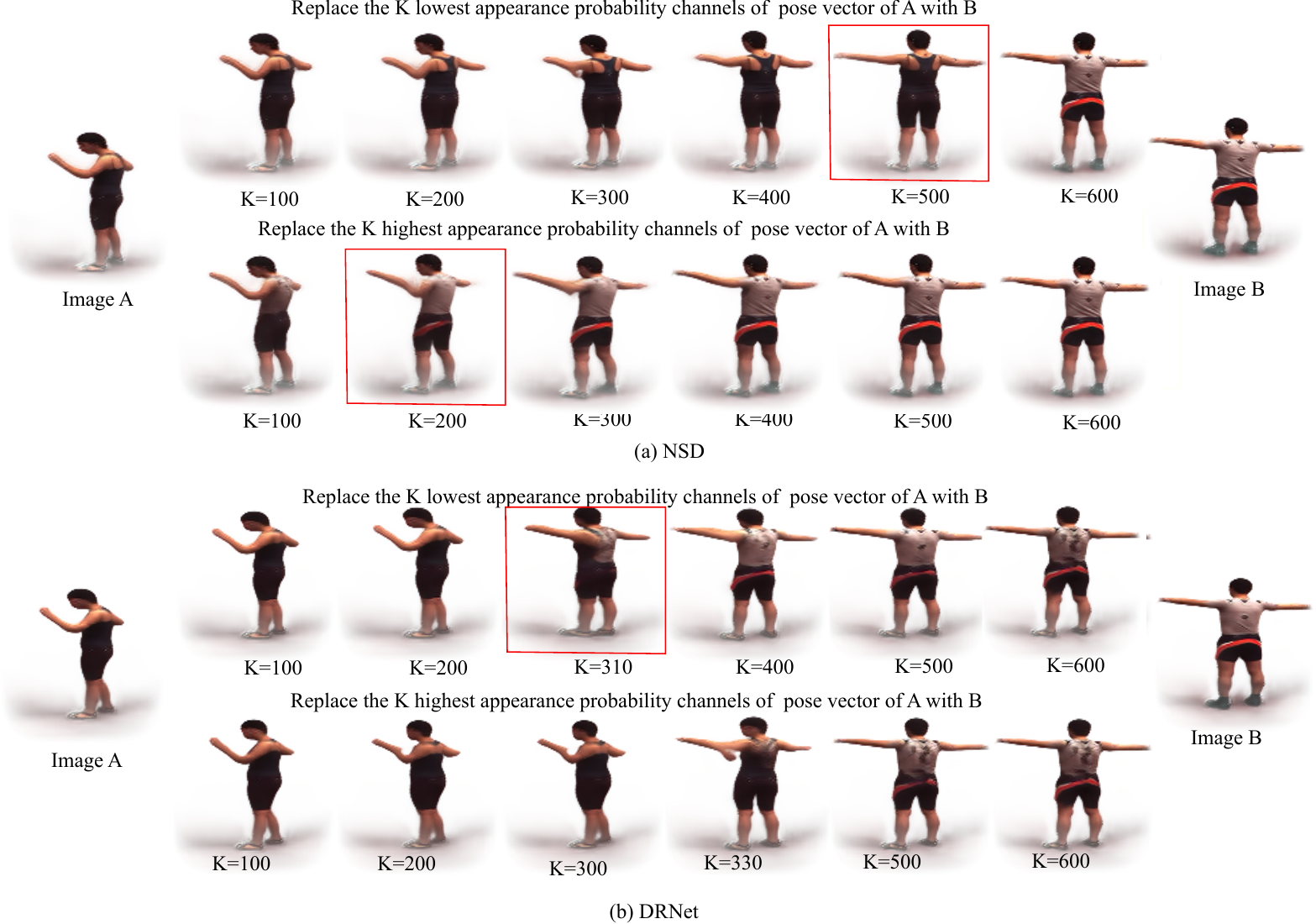}
    \end{tabular}
    \addtolength{\tabcolsep}{5pt}  %
    \end{center}
    \vspace{-6mm}
    \caption{ 	
    	{
    	{\bf Influence of the pose code channels.}  To synthesize the images in the middle portion of the figure, we take the appearance code corresponding to image A, and vary the pose code in two ways. Specifically, in the top (or bottom) portion of the figure, we replace the $K$ channels with lowest (or highest) appearance probability with the corresponding ones from the pose code extracted from image B. %
    	 (a) For NSD, replacing the $K=500$ lowest appearance probability channels yields an image (highlighted with a red box) depicting B's pose and A's appearance. Similarly, replacing the $K=200$ highest  appearance probability channels produces B's appearance and A's pose. (b) We observe similar trends for DRNet, although the separation of appearance and pose inside the pose code is not as clear as for NSD.}
    }
    \label{fig:analysis3_main}
    \vspace*{-0.3cm}
\end{figure}

In this section, we study the impact of the pose vector on the synthesized images and further provide evidence of the presence of appearance information in the pose code. To this end, we identify channels encoding appearance information in the pose code. Our approach is based on the idea that two images depicting different subjects in similar poses should ideally have similar latent pose codes. The channels that have large differences therefore indicate the presence of appearance information. 

To illustrate this, we use the two images shown in Figure~\ref{fig:analysis3_additional}(a) and plot the absolute difference between the corresponding pose codes obtained by NSD\footnote{{Similar plots for the other networks are provided in the supplementary material.}} 
in Figure~\ref{fig:analysis3_additional}(b), ordering the channels by the magnitude of the difference. The latent pose indeed disagree in many channels. We define the probability of a channel to encode appearance information to be proportional to the absolute pose vector difference for that channel. Below, we then analyze the effect of altering the $K$ channels with highest or lowest appearance probability.

To this end, we take two 
images A and B, as shown in the left and right ends of Figure~\ref{fig:analysis3_main}, fix the appearance code as that of A.  We then replace the channels with either the $K$ lowest or highest appearance probability in the pose code of A with the corresponding values from the pose code of B. Note that all disentangled networks have a pose code of dimension 600, and therefore $K=600$ means replacing all the channels of the pose vector.  

As shown in Figure~\ref{fig:analysis3_main}(a) for NSD, by replacing the $K=500$ lowest appearance probability channels yields an image (highlighted with a red box) with A's appearance and B's pose. Furthermore, replacing the $K=200$ highest appearance probability channels synthesizes an image with B's appearance and A's pose.  Both these results indicate that the top 100-200 highest probability appearance channels in the pose code indeed encode the appearance information for NSD.  It is worth noting that with $K=600$ the image depicts both the pose and appearance of B, confirming our previous experiments in Figure~\ref{fig:analysis1_main}. 

Figure~\ref{fig:analysis3_main}(b) for DRNet shows the channels are not as clearly separated in pose and appearance ones in this method. Nevertheless, the pose codes still combines pose and appearance information. We present similar analysis and visualizations for CSSL in the supplementary material.

\section{Disentanglement w.r.t. the 3D Pose Regressor}

The previous set of analyses have focused on the self-supervised representation learning networks themselves, evidencing that the latent pose vector is contaminated with appearance information. 
Here, we further investigate the disentanglement w.r.t. the supervised 3D human pose regressor, which takes the latent pose vector as input.  
Note that, since the 3D pose regressor is disassociated from the appearance vector at network level, studying the appearance and pose vector disentanglement in this context is not straightforward. Therefore, we consider the pose estimation network comprised of the self-supervised encoder and the supervised decoder as a standalone network and study the effects of the input image appearance on its 3D pose output.  
To this end, we introduce an adversarial perturbation strategy that explicitly focuses on modifying only the appearance information in the input image. Below, we first describe our attack framework, and then analyze its effects on the disentangled pose estimation networks.

\label{sec:attacks}
\vspace*{-0.3cm}

\subsection{Appearance-only Attack Framework}

 Our goal is to perturb only the subject's appearance in the input image; perturbing the image such that the subject's pose visually changes would of course make the pose regressor output a different pose but would not allow us to verify the disentanglement of pose and appearance. To enforce such a constraint on our perturbations, we follow a strategy that, intuitively, should constitute a weak attack and thus be favorable to the disentangled network. Specifically,  
we only perturb the latent appearance vector, which we combine with the \emph{original} pose one to generate an adversarial image. We then extract a new latent pose vector from this image and predict the 3D human pose from it. If the pose regressor could discard the appearance information, it would thus not be affected by this perturbation.

As shown in Algorithm~\ref{alg:attack}, we generate an adversarial image $\bI_{adv}$ as using a generator network $G$. In practice, we take $G$ to be either the disentangled network of interest or another disentangled network, and we will report results with both strategies.  First, we pass the original input image to the generator's spatial transformer $G_s$ and extract the cropped image $\bI_c$ using the resulting bounding box.  We then encode the cropped image $\bI_c$ into an initial latent pose vector $\tilde{\bh}_p^0$ and latent appearance vector $\tilde{\bh}_a^0$ using the generator's encoder $G_e$.  The combined latent vector $\tilde{\bh}=[\tilde{\bh}_a^0, \tilde{\bh}_p^0]$ is then passed as input to the generator's decoder $G_d$, which outputs the reconstructed image $\tilde{\bI}_{c}^0$ and a mask $\bM^0$. The cropped output $\tilde{\bI}_{c}^0$ is then combined with the pre-computed background image $\bB$ to resynthesize an image $\bI_{adv}^0$ at full resolution. This image then acts as input to the target pose estimation network, which encompasses an encoder $E$, that may differ from the generator one $G_e$, and a pose regressor. This forward pass produces an initial pose estimate $\phi(\bh_p^0)$. Note that the output of the target network given $\bI_{adv}^0$ as input has empirically a small mean per-joint position error  (MPJPE) of around 20 mm with respect to the prediction $\bq$ obtained from the original image $\bI$. This is because, at this point, no attack has been performed.

To attack only the subject's appearance in the adversarial input, we fix the pose vector $\tilde{\bh}_p=\tilde{\bh}_p^0$ to generate images of depicting the subject in their original pose. Furthermore, we also fix the mask to its initial value $\bM=\bM^0$. We then compute an appearance-only perturbation by optimizing the latent appearance vector $\tilde{\bh}_a$ in an iterative manner until it either achieves an MPJPE error with respect to the original prediction $\bq^0$ higher than a threshold, or reaches a maximum number of iterations.  {Note that our previous set of experiments in Section~\ref{sec:synthesis} have evidenced that modifying the appearance vector does not change the observed subject's pose, which validates our use of the network's decoder to generate the appearance-modified image.} %

\begin{algorithm}[!t]
	\begin{small}
		\caption{Appearance-only attacks} \label{alg:attack}
		\begin{algorithmic}[1]
			\Require \small $\bI$: Input image,  $G$: Pre-trained generator (with spatial transformer $G_s$, encoder $G_e$ and decoder $G_d$), $S$: Target spatial transformer, $E$: Target encoder, $D$: Target image decoder, $\phi$: Target pose regressor
			\State 			$\bI_{c} \gets G_t(\bI)$,
			$[\tilde{\bh}_{a}^{0}, \tilde{\bh}_p^0] \gets G_{e}(\bI_c)$
			\State $ \bI_{adv}^0 =  G_d(\tilde{\bh}_a^0, \tilde{\bh}_p^0), 	[\bh_a^0, \bh_p^0] \gets  E(S(\bI_{adv}^0))$
		\State	$  [\bh_a, \bh_p] \gets  E(S(\bI)),  $
						\State $\bq \gets  \phi(\bh_p)),  \textrm{error}_0 =  {\left\lVert \bq - \phi(\bh_p^0) \right \rVert}^2$
			\State 		$i \gets 1$
			\While   $\ \text{error}_i \leq \text{min. error}\  \text{ and } \ i \leq \text{max. iterations}$
			\State $\bI_{adv}^i  \gets  G_{d}(\tilde{\bh}_a^i, \tilde{\bh}_p^0)$
			\State $ [\bh_a^i, \bh_p^i] \gets  E(S(\bI_{adv}^i))$
			\State $\text{error}_i \gets   {\left\lVert  \bq - \phi(\bh_p^i)\right \rVert}^2$  \comment{\bq^0 - \phi(\bh_p^i)$}
			\State $\tilde{\bh}_{a}^{i+1} \gets \text{BackProp}\left\{ \text{error}_i  \right\}$ 
			\State $i \gets i+1$
			\EndWhile
			
			\State \Return $\bI_{adv} = \bI_{adv}^i$
		\end{algorithmic}
	\end{small}
\end{algorithm}

\vspace*{-0.4cm}
\subsection{Appearance-only  Attack Results}\label{sec:results}

\begin{figure}[!t]
    \begin{center}
    \begin{tabular}{cccc}
        \includegraphics[width=0.85\linewidth]{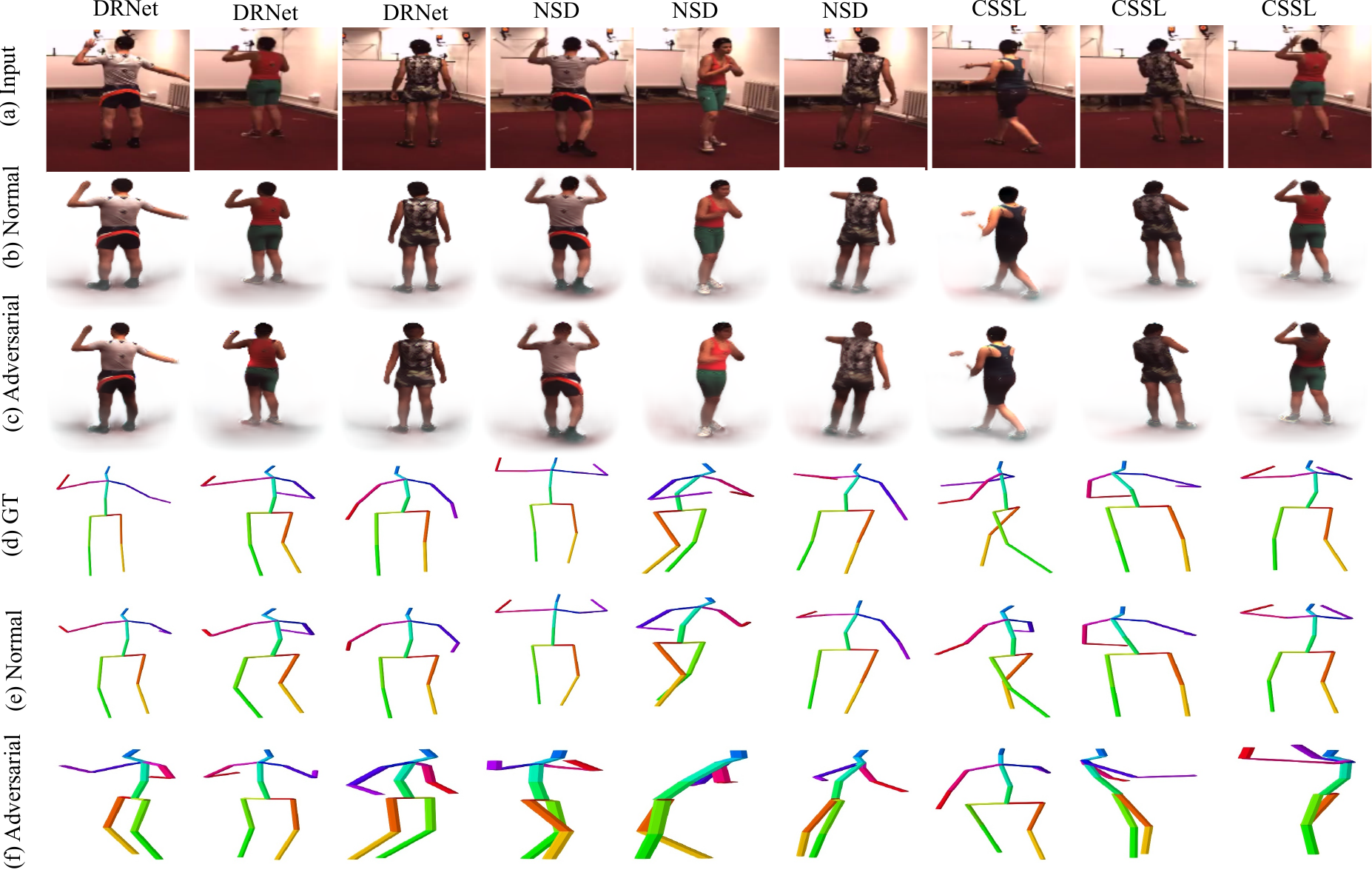}
    \end{tabular}
    \end{center}
   \vspace{-6mm}
    \caption{ 
    	{\bf Apperance-only Attack Examples.} Given an input image {\bf (a)}  with ground-truth pose {\bf (d)}, we first reconstruct  {\bf (b)}  the images using a generator.  By optimizing the latent appearance vector, we obtain an adversarial image {\bf (c)} that aims to fool the pose regressor so that it outputs a 3D pose {\bf (f)} that differs significantly from the original predictions {\bf (e)}. %
    }
    \label{fig:comparison_results}
    \vspace*{-0.7cm}
\end{figure}

\paragraph{\bf Qualitative Results.}
In Figure~\ref{fig:comparison_results}, we visualize the results of different models on the attacked images. For all disentangled representation frameworks, small changes in appearance produce wrong predictions.  In particular, as shown in  the third row, a small change in the shirt color leads to a completely different pose for all models. This demonstrates that the pose estimation network is dependent on {the subject's appearance in the input image} that its intermediate latent pose vector is not completely disentangled from appearance.

\paragraph{\bf Quantitative Study.}
We provide the results of our appearance-only attacks  in Table~\ref{tbl:attack} using the network decoder as the generator.  We report the MPJPE at the initial iteration and after the attack for each subject.  Specifically, the initial error corresponds to the error between the predictions obtained from the original image $\bI$ and from the synthesized image $\bI_{adv}^0$, without any latent attack. It is around 21.8 mm on average. This shows that the generator faithfully reconstructs the input image and can therefore be employed to perform the attack.  After the attack, the performance decrease across all the disentangled models
In other words, all models are vulnerable to our appearance-based attacks and typically reach an MPJPE of at least 175 mm. This indicates that the latent pose vector $\bh_p$ is not invariant to appearance changes and therefore that the appearance-pose disentanglement is not complete.  We provide ablative study using the same NSD decoder as the generator for all disentangled networks in the supplementary material.

{To further evaluate quantitatively the sensitivity of a disentangled network to our appearance-only attacks, we computed three image-based metrics, Peak Signal-to-Noise Ratio (PSNR), Structural Similarity Index (SSIM), and Mean Square Error (MSE), to compare the attacked images with those synthesized with the original framework. As shown in Table ~\ref{tbl:image_metrics_basic},
the three metrics indicate that the images obtained by attacking DRNet are more similar to the original synthesized ones than those obtained by attacking NSD or CSSL. This suggests that DRNet can be attacked with smaller changes, and thus contains more appearance information in its pose vectors.}

Altogether, our experiments evidence that disentangling pose and appearance in an unsupervised manner for 3D human pose estimation remains far from being solved. Our attacks thus provide a valuable testbed to valuate the effectiveness of future disentanglement-based frameworks.

\begin{table}[!ht]
	\centering
	
	\begin{minipage}{.52\textwidth}\centering
		\resizebox{\textwidth}{!}{
			\begin{tabular}{clccccccccccc}
				\toprule
				\multirow{2}{*}{\begin{tabular}[c]{@{}c@{}} {\bf Subject} \\  \end{tabular}}    		
				& \multicolumn{2}{c}{\bf NSD}
				& \multicolumn{2}{c}{\bf DRNet}   
				& \multicolumn{2}{c}{\bf CSSL}
				& \multicolumn{2}{c}{\bf Average}
				\T  \\			
				\cmidrule(l{5pt}r{5pt}){2-3} 
				\cmidrule(l{5pt}r{5pt}){4-5} 
				\cmidrule(l{5pt}r{5pt}){6-7} 
				\cmidrule(l{5pt}r{5pt}){8-9} 
				&  Initial & Final&Initial & Final& Initial & Final   & Initial & Final\T \\			
				\midrule
				{\bf 	S1}    & 21.0 & 179.7 & 23.9 &  169.7 & 21.5 & 176.9  & 21.6 & 174.2\T \\
				{\bf  S5}   & 19.6 & 180.0 & 14.1 & 166.7  & 25.3 & 186.5 & 19.6 & 177.1 \\
				{\bf 	S6 }  & 22.3 & 179.8 & 23.5 & 177.9 & 26.8 & 196.7& 23.4 & 184.7\\
				{\bf S7 }& 18.8 & 179.2 & 17.6 & 177.5 & 24.1 & 191.8 & 20.3 & 182.3\\
				{\bf 	S8 }& 16.8 & 178.6 & 21.7 & 198.9 &  30.5& 186.9 & 23.0 & 187.8\B \\			
				\hline
				{\bf Average }  & 	{\bf 19.7} & 	{\bf 179.5} & 	{\bf 20.2} & 	{\bf 177.5} & 	{\bf 25.6} & 	{\bf 207.5} & 	{\bf 21.8} & 	{\bf 176.8} \T \\
				\bottomrule
		\end{tabular}
	}
		\vspace{+0.05in}
		\caption{  {\bf  MPJPE before and after our appearance-based attacks.} We report the results of three networks and observe that disentangled networks 
			are  vulnerable to our attacks.	
	 }
		\label{tbl:attack}
	\end{minipage}\hfill
	\begin{minipage}{.43\linewidth}\centering
		\resizebox{0.60\columnwidth}{!}{
			\begin{tabular}{llccccccccccc}
				\toprule
				Metric
				& \multicolumn{1}{c}{ NSD}
				& \multicolumn{1}{c}{ DRNet}   
				& \multicolumn{1}{c}{CSSL}
				\T  \\
				\midrule
				SSIM$\uparrow$ & 0.947& 0.963 & 0.943  \\ %
				PSNR$\uparrow$ & 24.65 & 26.45 &24.37  \\%& 22.90\\
				MSE\ms{$\downarrow$} & 0.012 & 0.007 &  0.013\\ %
				\bottomrule
		\end{tabular}}
		\vspace{+0.05in}
		\caption{ Quantitative comparison of adversarial images with the original synthesized images. 
			These number show that the images obtained by attacking DRNet are closer to the original synthesized ones, and thus that the DRNet pose vectors tend to contain more appearance information.
		}
		\label{tbl:image_metrics_basic}
	\end{minipage}
\end{table}

\comment{
\begin{table}[!ht]
	\vspace*{-0.4cm}
	\centering
	
	\begin{minipage}{.52\textwidth}\centering
		\resizebox{\textwidth}{!}{
			\begin{tabular}{clccccccccccc}
				\toprule
				\multirow{2}{*}{\begin{tabular}[c]{@{}c@{}} {\bf Subject} \\  \end{tabular}}    		
				& \multicolumn{2}{c}{\bf AE}   
				& \multicolumn{2}{c}{\bf NSD}
				& \multicolumn{2}{c}{\bf DRNet}   
				& \multicolumn{2}{c}{\bf CSSL}
				& \multicolumn{2}{c}{\bf Average}
				\T  \\
				
				\cmidrule(l{5pt}r{5pt}){2-3} 
				\cmidrule(l{5pt}r{5pt}){4-5} 
				\cmidrule(l{5pt}r{5pt}){6-7} 
				\cmidrule(l{5pt}r{5pt}){8-9} 
				\cmidrule(l{5pt}r{5pt}){10-11} 
				&  Initial & Final&Initial & Final&Initial & Final& Initial & Final   & Initial & Final\T \\
				
				\midrule
				{\bf 	S1} &   23.5 & 178.8 & 21.0 & 179.7 & 23.9 &  169.7 & 21.5 & 176.9  & 20.8 & 176.2\T \\
				{\bf  S5} &  21.3  & 177.2 & 19.6 & 180.0 & 14.1 & 166.7  & 40.3 & 186.5 & 18.3 & 175.1 \\
				{\bf 	S6 }&  24.4 & 179.7 & 22.3 & 179.8 & 23.5 & 177.9 & 46.8 & 196.7& 21.4 & 178.7\\
				{\bf S7 }&  21.4 & 179.3 & 18.8 & 179.2 & 17.6 & 177.5 & 44.1 & 191.8 & 18.3 & 175.3\\
				{\bf 	S8 }&  18.0 & 179.4 & 16.8 & 178.6 & 21.7 & 198.9 &  30.5& 186.9 & 16.7 & 177.8\B \\
				
				\hline
				{\bf Average } & 	{\bf 21.7} & 	{\bf 178.9} & 	{\bf 19.7} & 	{\bf 179.5} & 	{\bf 15.3} & 	{\bf 171.5} & 	{\bf 19.6} & 	{\bf 177.5} & 	{\bf 19.1} & 	{\bf 176.8} \T \\

				\bottomrule
				
		\end{tabular}}
		\vspace{+0.05in}
		\caption{  {\bf  MPJPE before and after our appearance-based attacks.} We report the results of four networks and observe that disentangled networks are equally vulnerable to our attacks to the non-disentangled autoencoder. }
		\label{tbl:attack}
	\end{minipage}\hfill
	\begin{minipage}{.43\linewidth}\centering
		\resizebox{0.60\columnwidth}{!}{
			\begin{tabular}{llccccccccccc}
				\toprule
				Metric
				& \multicolumn{1}{c}{ NSD}
				& \multicolumn{1}{c}{ DRNet}   
				& \multicolumn{1}{c}{CSSL}
				\T  \\
				\midrule
				SSIM$\uparrow$ & 0.947& 0.963 & 0.943  \\ %
				PSNR$\uparrow$ & 24.65 & 26.45 &24.37  \\%& 22.90\\
				MSE\ms{$\downarrow$} & 0.012 & 0.007 &  0.013\\ %
				\bottomrule
		\end{tabular}}
		\vspace{+0.05in}
		\caption{ Quantitative comparison of adversarial images with the original synthesized images. The images obtained with DA are less similar to the original synthesized ones.}
		\label{tbl:image_metrics_basic}
	\end{minipage}
	\vspace*{-1cm}
\end{table}

}

\comment{

\begin{table}[t]
	\centering
	\resizebox{0.6\columnwidth}{!}{
		\begin{tabular}{clccccccccccc}
			\toprule
			\multirow{2}{*}{\begin{tabular}[c]{@{}c@{}} {\bf Subject} \\  \end{tabular}}    		
			& \multicolumn{2}{c}{\bf AE}   
			& \multicolumn{2}{c}{\bf NSD}
			& \multicolumn{2}{c}{\bf DRNet}   
			& \multicolumn{2}{c}{\bf CSSL}
			& \multicolumn{2}{c}{\bf Average}
			  \T  \\
			
			\cmidrule(l{5pt}r{5pt}){2-3} 
			\cmidrule(l{5pt}r{5pt}){4-5} 
			\cmidrule(l{5pt}r{5pt}){6-7} 
			\cmidrule(l{5pt}r{5pt}){8-9} 
						\cmidrule(l{5pt}r{5pt}){10-11} 
		&  Initial & Final&Initial & Final&Initial & Final& Initial & Final   & Initial & Final\T \\
		
		\midrule
	{\bf 	S1} &   23.5 & 178.8 & 21.0 & 179.7 & 17.1 &  169.5 & 21.5 & 176.9  & 20.8 & 176.2\T \\
			{\bf  S5} &  21.3  & 177.2 & 19.6 & 180.0 & 14.1 & 166.7  & 18.2 & 176.5 & 18.3 & 175.1 \\
		{\bf 	S6 }&  24.4 & 179.7 & 22.3 & 179.8 & 16.1 & 176.1 & 22.7 & 179.4& 21.4 & 178.7\\
			{\bf S7 }&  21.4 & 179.3 & 18.8 & 179.2 & 14.8 & 166.1 & 18.4 & 176.8 & 18.3 & 175.3\\
		{\bf 	S8 }&  18.0 & 179.4 & 16.8 & 178.6 & 14.4 & 179.2 &  17.4& 177.9 & 16.7 & 177.8\B \\
		
		\hline
			{\bf Average } & 	{\bf 21.7} & 	{\bf 178.9} & 	{\bf 19.7} & 	{\bf 179.5} & 	{\bf 15.3} & 	{\bf 171.5} & 	{\bf 19.6} & 	{\bf 177.5} & 	{\bf 19.1} & 	{\bf 176.8} \T \\

		\bottomrule
		
	\end{tabular}}
	\vspace{+0.05in}
	\caption{  {\bf  MPJPE before and after our appearance-based attacks.} We report the results of four networks and observe that disentangled networks are equally vulnerable to our attacks to the non-disentangled autoencoder. \KN{to be updated.}}
	\label{tbl:attack}
\end{table}

}

{
}

\vspace*{-0.5cm}

\section{Discussion}

\paragraph{\bf Evaluating Disentanglement.} Several methods~\cite{eastwood2018framework,do2019theory,liu2021measuring} have been proposed for assessing the degree of disentanglement of latent variables. In particular, we report the two complementary state-of-the-art metrics of~\cite{liu2021measuring}, Distance Correlation (DC)  and Information over Bias (IOB) to evaluate disentanglement.  DC is bounded in [0,1] and measures the correlation between the two latent spaces; IoB measures the amount of information from the input image that is encoded in a given latent space.
In Table~\ref{tbl:metrics}, we provide these metrics, averaged over 400 images, for the pose (P) and appearance (A) latent spaces and for different disentanglement strategies.  DC(A, P) contain large values indicating that the appearance and pose are correlated.  Furthermore, the IOB(I, P) values are larger than the IOB(I, A), which suggests that the pose code encodes more input information than the appearance code.  Note that DC(A, P) cannot be  used as a standalone metric to interpret disentanglement because low values of DC can also indicate noise in one latent space.  While DRNet achieves the best DC(A, P) score, its value of 0.90  IOB(I, A) suggests that the appearance code encodes minimal information.  Although these metrics quantify disentanglement,
they offer little understanding of the disentanglement issues, and IOB is difficult to interpret because it is unbounded and requires training an external decoder network whose optimal architecture is unknown. By contrast, our analyses enable a finer-grain understanding of the pose and appearance latent spaces of representation learning strategies for human pose estimation, and provide visual results that are easier to interpret.

    \begin{table}[!ht]
    \centering
\begin{minipage}{.49\linewidth}\centering
			\resizebox{0.65\columnwidth}{!}{
		\begin{tabular}{llccc}
		\toprule
		Metric
		& \multicolumn{1}{c}{ NSD}
		& \multicolumn{1}{c}{ DRNet}   
		& \multicolumn{1}{c}{CSSL}
		\T  \\
		\midrule
		DC(A, P)$\downarrow$ & 0.88  & 0.59 & 0.77\\ %

		IOB(I, A)$\uparrow$ & 0.79 & 0.90 & 0.95\\ %
		IOB(I, P)$\uparrow$ & 1.15 & 1.08 & 1.29 \\ %
		\bottomrule
\end{tabular}}
	\vspace{+0.05in}
\caption{ {  \small Disentanglement-related metrics for the pose (P) and appearance (A) latent spaces extracted from an input image (I).
	} }
\label{tbl:metrics}
\end{minipage}\hfill
\begin{minipage}{.49\linewidth}\centering
					\resizebox{0.5\columnwidth}{!}{
		\begin{tabular}{llccccccccccc}
		\toprule
		Metric
		& \multicolumn{1}{c}{CSSL}
		& \multicolumn{1}{c}{ CSSL(DA)}
		\T  \\
		\midrule
		SSIM$\uparrow$ & 0.943   & 0.926 \\
		PSNR$\uparrow$  &24.37  & 22.90\\
		MSE\ms{$\downarrow$} &  0.013&  0.018\\
		\bottomrule
\end{tabular}}
	\vspace{+0.05in}
\caption{ Quantitative comparison of adversarial images with  original synthesized images. The images obtained with DA are less similar to  original synthesized ones.
}
\label{tbl:image_metrics}
\end{minipage}

\end{table}

\comment{
\begin{table}[t]
	\centering
	\resizebox{0.45\columnwidth}{!}{
		\begin{tabular}{clccccccccccc}
			\toprule
			Metric
			& \multicolumn{1}{c}{ NSD}
			& \multicolumn{1}{c}{ DRNet}   
			& \multicolumn{1}{c}{CSSL}
			& \multicolumn{1}{c}{ CSSL(DA)}
			\T  \\
			
			\midrule
			
			DC(A, P)$\downarrow$ & 0.88  & 0.59 & 0.77 & 0.62 \\
			DC(I, A)$\uparrow$ & 0.82 & 0.90 &0.89 & 0.66\\
			DC(I, P)$\uparrow$ & 0.85 & 0.61 & 0.67 & 0.84\\
			IOB(I, A)$\uparrow$ & 0.79 & 0.90 & 0.95 & 1.07 \\
			IOB(I, P)$\uparrow$ & 1.15 & 1.08 & 1.29 & 1.21\\
			
			\bottomrule
			
	\end{tabular}}
	\vspace{+0.05in}
	\caption{ {   Disentanglement, informativness metrics for pose and appearance  latent space.} }
	\label{tbl:metrics}
\end{table}

}

\comment{
\begin{figure}[!t]
    \begin{center}
    \addtolength{\tabcolsep}{-5pt}  %
    \begin{tabular}{cccc}
    \includegraphics[width=\linewidth]{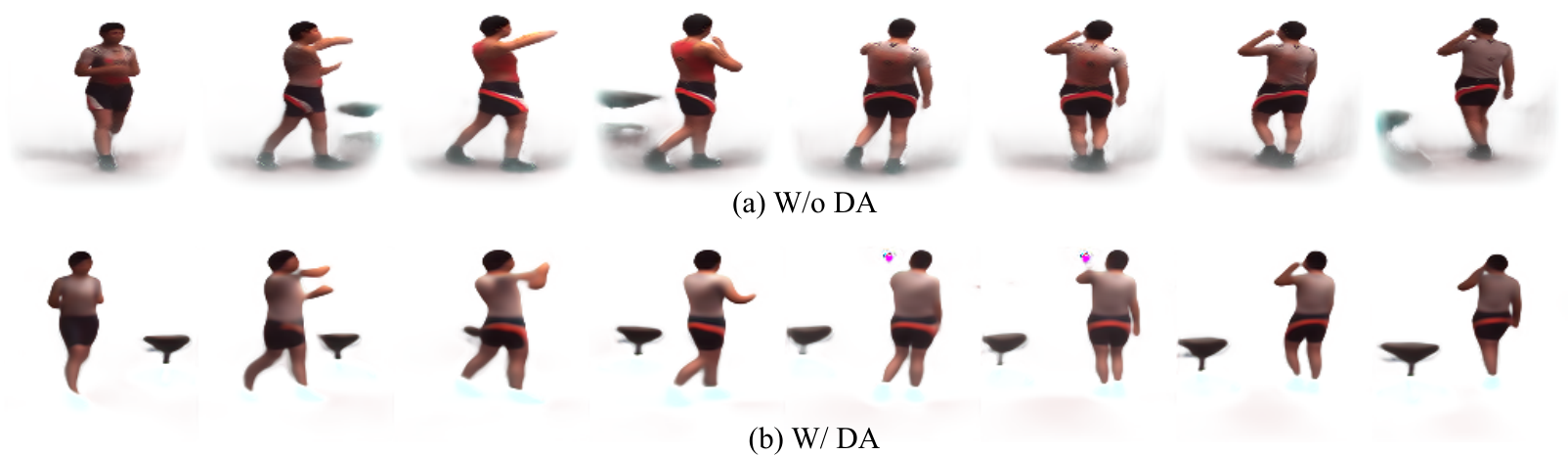}
    \end{tabular}
    \addtolength{\tabcolsep}{5pt}  %
    \end{center}
    \vspace{-6mm}
    \caption{ 	
    	 {\bf Synthesizing novel images with CSSL (DA).} As in Figure~\ref{fig:analysis1}, we take S7's pose vector and S8's appearance one and synthesize novel images with CSSL, either without (top) or with (bottom) DA during training. The image synthesized with CSSL (DA)  retain S8's appearance without residual red shirt color from S7.}
    \label{fig:data_aug_novel_syn}
\end{figure}

}

\begin{figure}
	\centering
			\resizebox{0.5\textwidth}{!}{
		\begin{minipage}{.45\textwidth}
		\centering
        \includegraphics[width=1.0\linewidth]{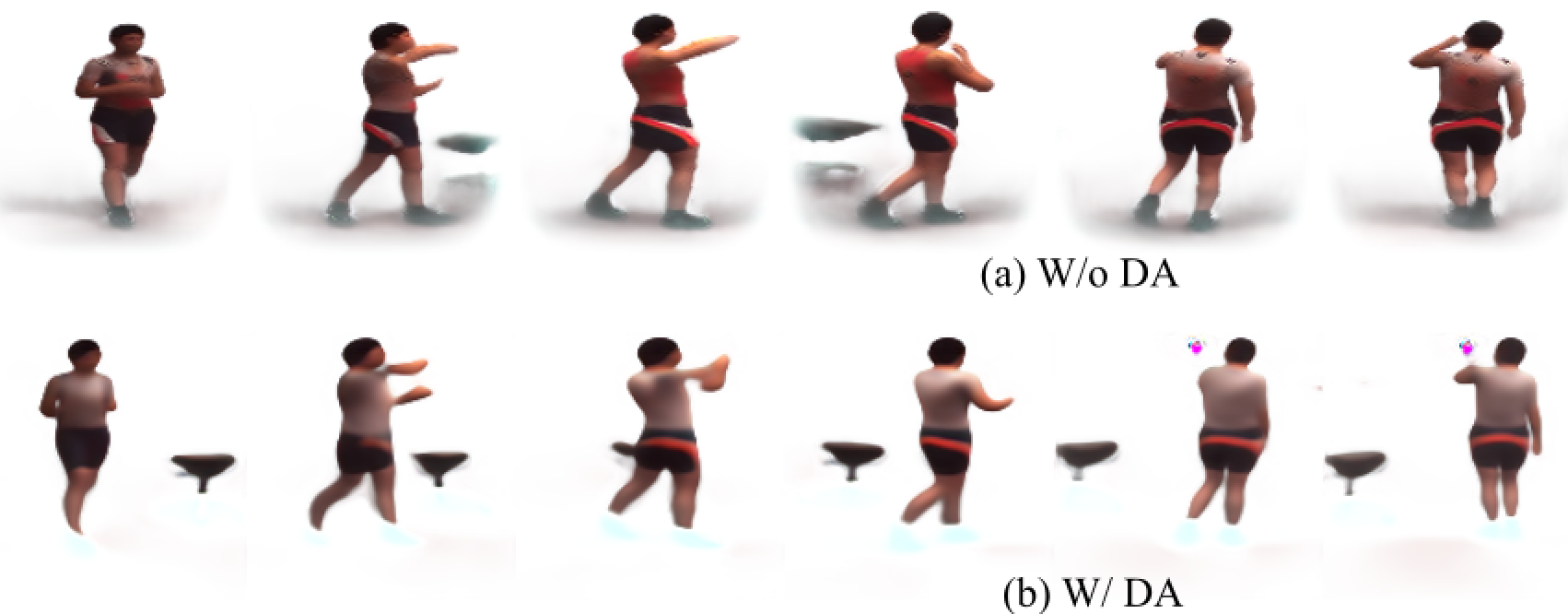}
\caption{ 	
	{\bf Synthesizing novel images with CSSL (DA).} As in Figure~\ref{fig:analysis1_main}, we take S7's pose vector and S8's appearance one and synthesize novel images with CSSL, either without (top) or with (bottom) DA during training. The image synthesized with CSSL (DA)  retain S8's appearance without residual red shirt color from S7.}
    \label{fig:data_aug_novel_syn}
	\end{minipage} \hfill
}
			\resizebox{0.4\textwidth}{!}{
	\begin{minipage}{.45\textwidth}
		\centering
    \includegraphics[width=0.8\linewidth]{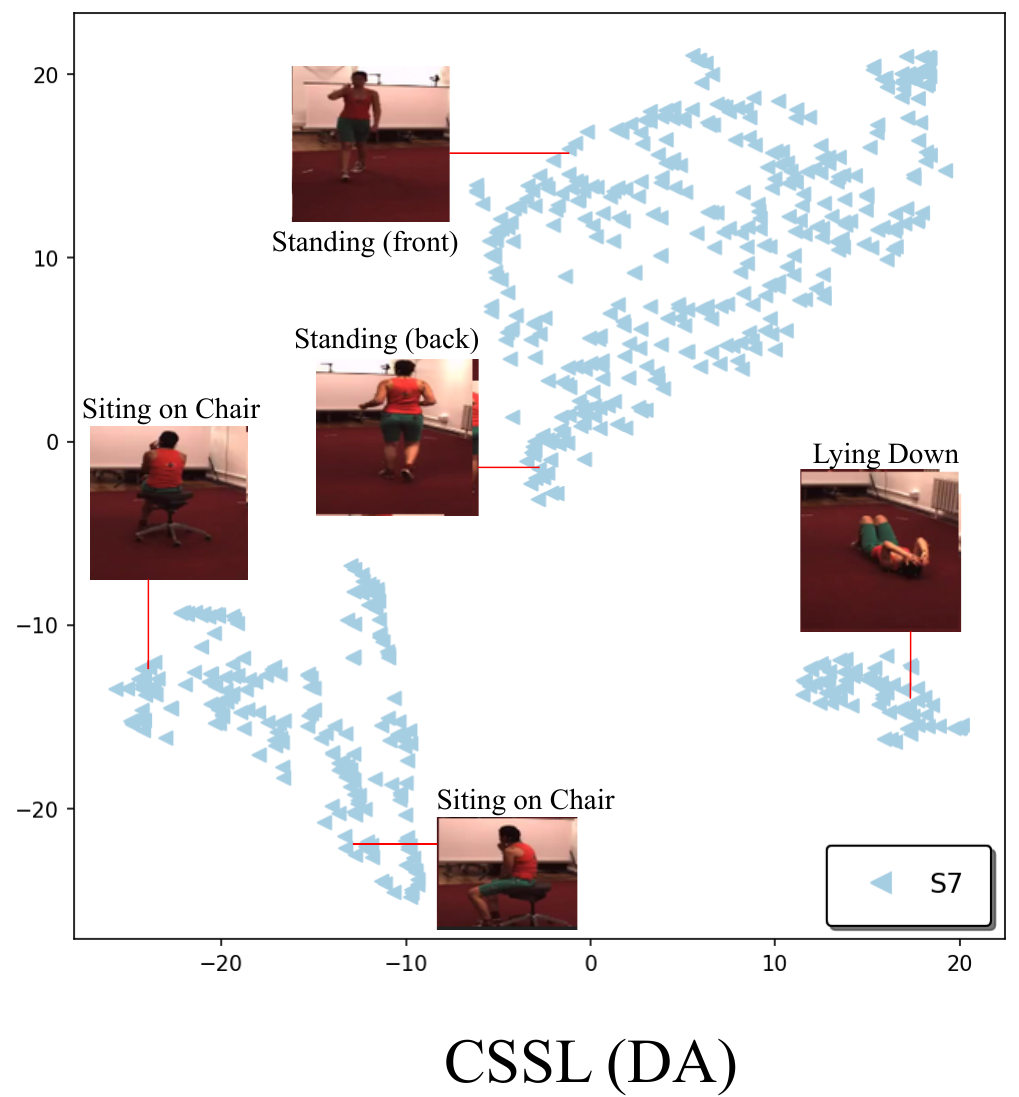}
	  \caption{ 
		{\bf tSNE visualization of  CSSL (DA) appearance codes.} The appearance codes of images from same subject are  stilll clustered according to the action performed by the subject. %
	}
	\label{fig:tsne_cssl_DA}
	\end{minipage}
}
\end{figure}

\begin{figure}[!t]
    \begin{center}

    \includegraphics[width=0.8\linewidth]{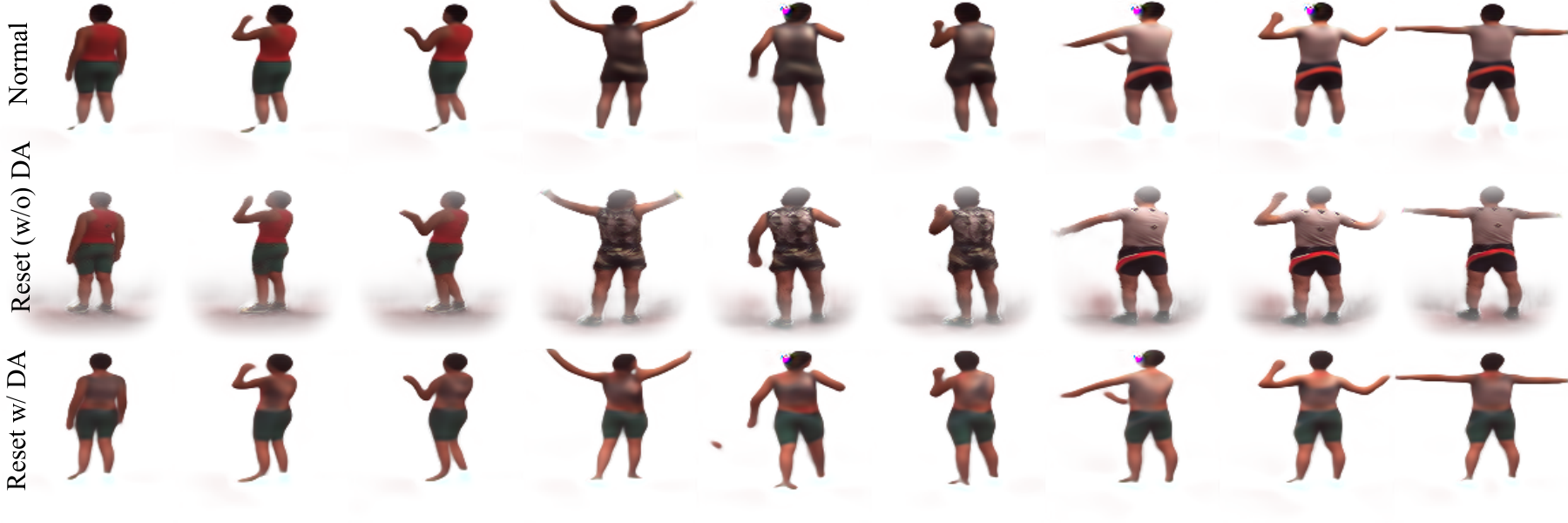} 
    \end{center}
   \vspace{-6mm}
    \caption{ {\bf Zero appearance vectors with CSSL (DA).}
In first row, we show the original image synthesized with CSSL. While, without DA (middle), the synthesized images obtained with a zero appearance vector retain the original subject's appearance, with DA (bottom), all the subjects have a similar the appearance. This suggests that DA helps to remove appearance information from the pose vectors.
  }
    \label{fig:data_augmentation}
    \vspace*{-5mm}
\end{figure}

\paragraph{\bf Does data-augmentation help to learn appearance-invariant features?}
Recently, powerful data augmentation (DA) strategies, such as AugMix~\cite{augmix}, CutMix~\cite{cutmix} and others~\cite{hong2021stylemix,zhang2018mixup}, have been proposed to improve the generalization power and robustness of neural networks. Furthermore, classical adversarial training~\cite{madry2018towards,kannan2018adversarial} can be viewed as a form of data augmentation with adversarial images. Here, we therefore study if data augmentation constitutes a promising direction towards more effectively disentangling self-supervised 3D human pose estimation networks.

Since the network architectures we consider are much more complicated than the image recognition ones used in the above-mentioned DA works, we employ a simpler DA strategy consisting of augmenting the output of the spatial transformer with RGB jitter. We then re-run the analyses we presented before, focusing here on CSSL. %
Specifically, in Figure~\ref{fig:data_aug_novel_syn}, we show the images synthesized when mixing S7's pose vector with S8's appearance. Note that, with DA, the images better retain the appearance of S8. Furthermore, in Figure~\ref{fig:data_augmentation}, we show images obtained by making use of a zero appearance vector. With DA, all the synthesized images depict a similar subject appearance. Altogether, this suggests that DA helps the disentanglement process in CSSL, which is further confirmed by the DC(A, P) value that improves from 0.77 to 0.62. This value of 0.62 nonetheless still indicates a relatively high correlation between the latent spaces. To further analyze this, {we computed a similar t-SNE plot as that of Figure~\ref{fig:tsne_cssl_DA}%
, and observed that the actions are still clustered, evidencing that the appearance code still contains some pose information.}

Similarly, we also ran our appearance-only attacks on the CSSL model trained with DA, and observed the attacks to remain successful, suggesting that the pose vector remains contaminated by appearance information. {To evaluate quantitatively whether DA nonetheless improved this, we report the PSNR, SSIM, and MSE metrics between the attacked images and the original synthesized ones in Table~\ref{tbl:image_metrics}. 
The values indicate that the images obtained by attacking the network without DA are more similar to the original synthesized ones. In other words, CSSL (DA) requires larger changes in the input image to attack the 3D pose regressor.}
Altogether, these results indicate that DA constitutes a promising direction to improve disentanglement, and we leave the development of more effective DA strategies as future work.

\section{Conclusion}
In this work, we have analyzed the latent vectors extracted by self-supervised disentangled networks for 3D human pose estimation. Specifically, we have studied the disentanglement of pose and appearance from the perspective of both the representation learning network, and the supervised 3D human pose regressor. In the former case, our analyses via diverse image synthesis strategies have evidenced that the state-of-the-art disentanglement-based representation learning networks do not truly disentangle pose from appearance, and in particular that the latent pose codes contain significant appearance information. In the latter, we have shown that disentanglement-based networks were \comment{no more}  \kn{not} robust to appearance-only adversarial attacks, 
despite these attacks being designed to be as favorable as possible to the disentanglement-based frameworks.
We believe that our analysis methodology and our semantic attacks will be beneficial to improve disentanglement-based representation learning in the future, and thus positively impact self-supervised 3D human pose estimation.

\bibliographystyle{splncs04}
\bibliography{egbib}
\end{document}